\documentclass[11pt]{article}

\usepackage[final]{acl}

\usepackage{times}
\usepackage{latexsym}

\usepackage[T1]{fontenc}

\usepackage[utf8]{inputenc}

\usepackage{microtype}

\usepackage{inconsolata}

\usepackage{graphicx}

\usepackage{amssymb}
\usepackage{multirow}
\usepackage{makecell}
\usepackage{amsmath}
\usepackage{booktabs}
\usepackage{diagbox}
\usepackage{pifont}
\usepackage[most]{tcolorbox}

\title{Beyond Entity Recognition: \\Vision-Centric Knowledge Visual Question Answering}

\author{
 \textbf{Jiatong Ma\textsuperscript{1,2,3}\thanks{Equal Contribution.}},
 \textbf{Yuchen Liu\textsuperscript{1,2}\footnotemark[1]},
 \textbf{Longteng Guo\textsuperscript{1,2}\footnotemark[1]},
 \textbf{Yichen Yuan\textsuperscript{1,2}},
 \textbf{Yuze Wu\textsuperscript{1,2}},\\
 \textbf{Dongze Hao\textsuperscript{3}},
 \textbf{Ru zhen\textsuperscript{3}}, 
 \textbf{Yanhao Zhang\textsuperscript{3}},
 \textbf{Haonan Lu\textsuperscript{3}},
 \textbf{Jing Liu\textsuperscript{1,2}\thanks{Corresponding author.}}
\\
\\
 \textsuperscript{1}
Institute of Automation, Chinese Academy of Sciences,
\\
 \textsuperscript{2}School of Artificial Intelligence, University of Chinese Academy of Sciences, 
\\
 \textsuperscript{3}OPPO AI Center, OPPO Inc. , 
\\
   \texttt{majiatong2025@ia.ac.cn, jliu@nlpr.ia.ac.cn}
}

\begin{document}
\maketitle

\begin{abstract}

We propose VC-VQA, a large-scale benchmark for vision-centric knowledge reasoning, aiming to systematically evaluate the capabilities of Multimodal Large Language Models (MLLMs) in fine-grained visual knowledge acquisition and complex multi-stage reasoning. Unlike existing benchmarks that primarily treat visual information as entity recognition and rely on text-based external knowledge, VC-VQA explicitly models the role of visual information by considering both its source (internal vs. external images) and stage of involvement (pre-retrieval vs. post-retrieval). Specifically, VC-VQA requires models to retrieve external visual evidence, identify visual clues before retrieval, and localize, verify, or match retrieved knowledge against the internal image. We further construct a controlled multimodal knowledge base with traceable textual and visual evidence for each sample. We evaluate 21 representative MLLMs under four settings: no external knowledge, with provided evidence, knowledge base retrieval, and open-web search. Experimental results lead to three key findings: current models still struggle with vision-centric knowledge reasoning; fine-grained visual knowledge acquisition remains the primary bottleneck; and agentic retrieval highlights the value of multi-step planning and cross-modal knowledge integration. Overall, VC-VQA provides a more challenging benchmark for evaluating vision-centric knowledge reasoning in MLLMs. Our code and dataset are available at \url{https://github.com/CASIA-IVA-Lab/VC-VQA}.

\end{abstract}


\section{Introduction}
With the rapid advancement of Multimodal Large Language Models (MLLMs), knowledge-based visual question answering has become an important benchmark for evaluating multimodal understanding and reasoning. While recent MLLMs have achieved remarkable progress in fine-grained visual recognition and textual knowledge acquisition, real-world questions require models to reason over visual knowledge throughout the entire reasoning process. For example, a question may require the model to extract visual cues from the input image for knowledge retrieval, retrieve additional external images, and jointly reason over both the retrieved images and the input image before reaching the final answer. Such tasks require visual information to play an active role beyond entity recognition.

\begin{figure*}
  \centering
  \includegraphics[width=\textwidth]{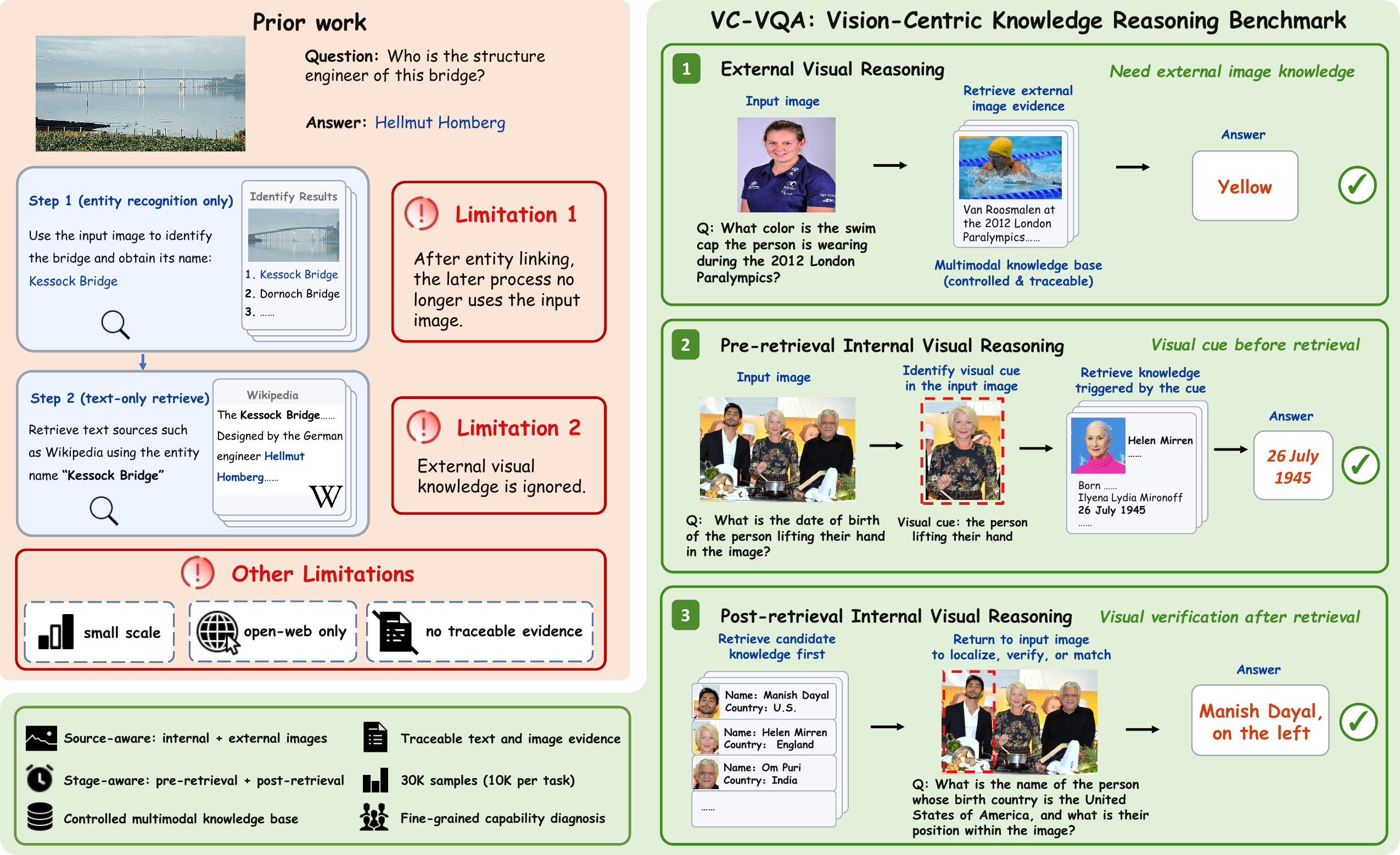}
  \caption{Comparison between previous work and our benchmark.}
  \label{comparison}
\vspace{-5pt}
\end{figure*}

Existing benchmarks, such as OK-VQA~\cite{marino2019ok}, A-OKVQA~\cite{schwenk2022okvqa}, KVQA~\cite{shah2019kvqa}, EVQA~\cite{mensink2023encyclopedic}, InfoSeek~\cite{chen2023can}, DynVQA~\cite{li2024benchmarking}, and M$^3$-VQA~\cite{ma2026m3}, have progressively increased reasoning complexity. However, these benchmarks still lack a systematic characterization of the role of visual information in knowledge reasoning. First, the input image is only used for entity recognition and is discarded after entity linking. In many real-world scenarios, however, visual information serves purposes beyond entity recognition. It may provide retrieval cues before knowledge retrieval or participate in localization, verification, and matching after retrieval. Second, external knowledge is almost exclusively represented as text, despite the fact that many real-world facts are inherently visual, such as the figure's clothing in a specific event, an appearance of a building during different historical periods. Recent benchmarks MM-BrowseComp~\cite{li2025mm} and VisBrowse-Bench~\cite{zhang2026visbrowse} have begun exploring external visual knowledge. However, they contain only a few hundred samples and rely on open-web search. As a result, the knowledge sources are dynamically changing, evidence is not traceable, experimental results lack robustness, and fine-grained capability diagnosis is difficult.

To address these limitations, we propose VC-VQA, a large-scale benchmark for vision-centric multimodal knowledge reasoning. VC-VQA is the first benchmark that systematically characterizes visual information according to both its source (input image vs.\ external image) and its role in reasoning (before vs.\ after retrieval). Accordingly, we define three representative reasoning tasks: External Visual Reasoning, which requires acquiring visual knowledge from external images in the knowledge base; Pre-retrieval Internal Visual Reasoning, which requires extracting visual information from the input image as retrieval cues before knowledge retrieval; and Post-retrieval Internal Visual Reasoning, which requires revisiting the input image after retrieval for localization, verification, and matching. Together, these tasks provide a systematic evaluation of multimodal reasoning under different visual knowledge paradigms. Furthermore, VC-VQA is built upon a controlled multimodal knowledge base constructed from a Wikipedia snapshot, with gold textual and visual evidence annotated for every sample, enabling stable, reproducible, and diagnostically informative evaluation.



\begin{table*}
  
  \centering
\resizebox{\linewidth}{!}{
  \small
  \begin{tabular}{lcccccccccccc}
\toprule
\textbf{}

& \textbf{OKVQA}
& \textbf{KVQA} 
& \textbf{InfoSeek} 
& \textbf{EVQA} 
& \textbf{Dyn-VQA} 
& \textbf{M$^3$-VQA } 
& \textbf{MM-BrowseComp} 
& \textbf{VC-VQA} 
\\

\midrule

Fine-Grained Entity & $\times$ & $\checkmark$ & $\checkmark$ & $\checkmark$ & $\checkmark$ & $\checkmark$ & $\checkmark$ & $\checkmark$ \\


Multi-Hop & $\times$ & $\checkmark$ & $\times$ & $\checkmark$ & $\checkmark$ & $\checkmark$ & $\checkmark$ & $\checkmark$ \\

Visual Reasoning Stage\\

~~~~$\hookrightarrow$Pre-retrieval & $\times$ & $\times$ & $\times$ & $\times$ & $\times$ & $\times$ & $\times$ & $\checkmark$ \\

~~~~$\hookrightarrow$Post-retrieval & $\times$ & $\times$ & $\times$ & $\times$ & $\times$ & $\times$ & $\times$ & $\checkmark$ \\

Visual Information Source\\

~~~~$\hookrightarrow$Internal Image & $\times$ & $\times$ & $\times$ & $\times$ & $\times$ & $\times$ & $\times$ & $\checkmark$ \\

~~~~$\hookrightarrow$External Image & $\times$ & $\times$ & $\times$ & $\times$ & $\times$ & $\times$ & $\checkmark$ & $\checkmark$ \\

Controlled Knowledge Base\\
~~~~$\hookrightarrow$Traceable Answer & $\times$ & $\times$ & $\times$ & $\checkmark$ & $\times$ & $\checkmark$ & $\times$ & $\checkmark$ \\

~~~~$\hookrightarrow$Free-Form KB & $\times$ & $\times$ & $\checkmark$ & $\checkmark$ & $\times$ & $\checkmark$ & $\times$ & $\checkmark$ \\


\midrule
Answer Type & Open & Open & Open & Open & Open & Open & Open & Open \\
\#\{I, Q, A\} & 14K & 183K & 1M & 1M & 1K & 13K & 0.2K & 30K \\


\bottomrule
\end{tabular}
}
\caption{Comparison between our VC-VQA and previous VQA datasets.}
\vspace{-5pt}
\label{datasets_comparison}
\end{table*}

Using VC-VQA, we evaluate a wide range of closed-source and open-source MLLMs under four settings: without external knowledge, with provided evidence, with retrieval from a controlled knowledge base, and with search from open-web. Our experiments lead to three key findings: 

\begin{itemize}
\item \textbf{Current MLLMs still struggle with vision-centric knowledge reasoning.} Even the strongest state-of-the-art model achieves an overall accuracy of only 57.1\% , while accuracy exceeds 90\% when gold evidence is provided. This large gap indicates that visual knowledge reasoning remains a major bottleneck for modern MLLMs.
\vspace{-3pt}
\item \textbf{Fine-grained visual knowledge acquisition remains the primary bottleneck.} Providing entity names substantially improves internal visual reasoning, whereas external visual reasoning still depends heavily on image evidence, which suggests that current models can effectively recall textual knowledge once an entity is identified, but still struggle to acquire fine-grained visual knowledge.
\vspace{-3pt}
\item \textbf{Agentic retrieval consistently outperforms heuristic retrieval.} Although retrieval augmentation consistently improves overall performance, agent-based retrieval achieves superior results across all settings. This demonstrates the importance of active planning, iterative retrieval, and coordinated acquisition of visual and textual knowledge for complex multimodal knowledge reasoning.

\end{itemize}

\section{Related Work}

With the advancement of MLLMs, research on multimodal knowledge reasoning has attracted increasing attention. Existing knowledge-based VQA benchmarks, such as OKVQA~\cite{marino2019ok}, A-OKVQA~\cite{schwenk2022okvqa}, have promoted the development of open-world visual question answering by incorporating external knowledge. Recent benchmarks such as KVQA~\cite{shah2019kvqa}, S3VQA~\cite{jain2021select}, ViQuAE~\cite{lerner2022viquae}, InfoSeek~\cite{chen2023can}, EVQA~\cite{mensink2023encyclopedic}, DynVQA~\cite{li2024benchmarking}, and M\textsuperscript{3}-VQA~\cite{ma2026m3} cover fine-grained entity attributes and have evolved toward more complex settings involving multi-hop reasoning, multiple entities, and diverse domains. However, their visual understanding remains shallow: images are only used in the initial stage to identify entities, and once entity linking is completed, visual information is entirely excluded from the reasoning process. Moreover, external knowledge is represented solely in textual form, overlooking the fact that a substantial portion of knowledge is inherently conveyed through images. More recently, benchmarks such as MM-BrowseComp~\cite{li2025mm} and VisBrowse-Bench~\cite{zhang2026visbrowse} have introduced questions that require external visual knowledge. However, they still suffer from three major limitations. First, they remain limited in scale (containing only 224 and 169 samples, respectively), making comprehensive evaluation difficult. Second, they rely on the open Internet as the knowledge source, where webpages and search results continuously change, resulting in poor reproducibility. Third, they focus primarily on whether a model produces the correct final answer, without providing traceable gold evidence or fine-grained capability categorization, making it difficult to identify where reasoning failures occur.

\section{Task Definition of VC-VQA}

The VC-VQA task can be formally defined as: $f:(I, Q, K) \rightarrow A$. where \(I\) denotes a single input image that may contain one or more visual entities; \(Q\) denotes a question querying fine-grained textual or visual knowledge; \(K\) denotes a multimodal knowledge base consisting of Wikipedia-style pages containing both text and images; and \(A\) denotes the answer. Each sample is accompanied by a set of gold evidence \(E\), consisting of one or more external textual or visual evidence items.

Based on the reasoning patterns required by the question \(Q\) and the corresponding evidence \(E\), we divide VC-VQA into three sub-tasks:
    \textbf{(1) External Visual Reasoning (Ex): }  
    Answering the question necessarily requires external image evidence from the knowledge base; textual knowledge alone is insufficient to derive the correct answer. This setting evaluates a model's ability to retrieve and utilize multimodal external knowledge, where the knowledge itself is primarily conveyed through images.
    \textbf{(2) Pre-Retrieval Internal Visual Reasoning (In-pre):}  
    The model must first extract the required visual information from the input image and then use it as a clue to retrieve relevant external knowledge before answering the question. This process goes beyond recognizing a single entity: visual attributes are used to identify entities, while visual relationships serve as triggers for knowledge retrieval.
    \textbf{(3) Post-Retrieval Internal Visual Reasoning (In-post): }  
    The model first retrieves candidate entity information through textual knowledge and subsequently returns to the input image for localization, verification, or matching before producing the final answer. In this setting, the input image remains actively involved in the reasoning process after knowledge retrieval, enabling a reverse mapping from knowledge to image.

\section{The VC-VQA Dataset}

\subsection{Dataset Construction Pipeline}

\begin{figure*}
  \centering
  \includegraphics[width=0.95\textwidth]{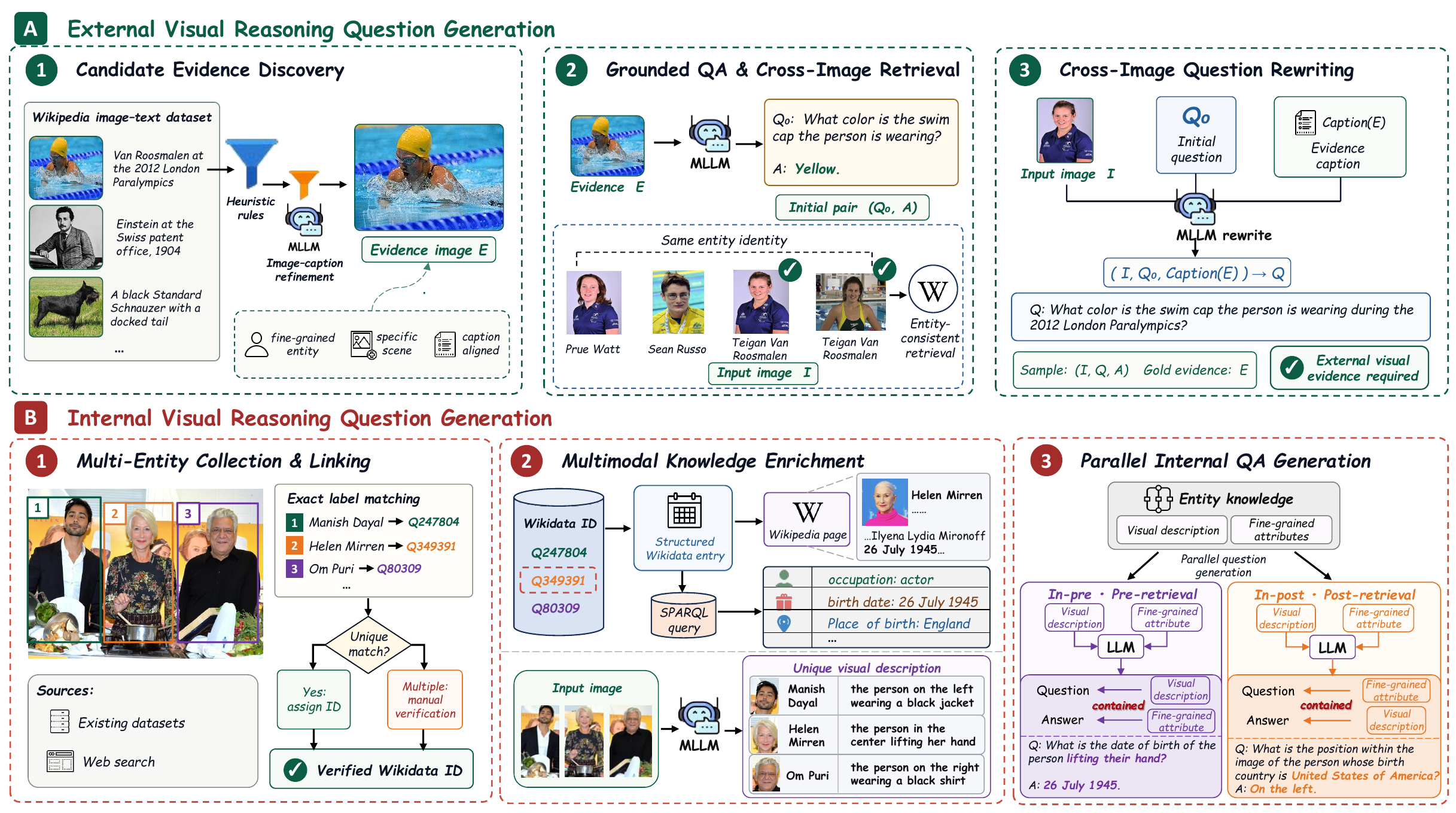}
  \caption{Data Construction Pipeline. See Appendix~\ref{appx:dataset} for more.}
  \label{comparison}
\vspace{-5pt}
\end{figure*}

\paragraph{External Visual Reasoning Question Generation.}
We construct external visual reasoning samples based on the large-scale Wikipedia image--text dataset WIT~\cite{srinivasan2021wit}. Given an image and its associated caption, we first employ heuristic rules and MLLMs to identify images containing fine-grained entities in specific contexts, which are then selected as external evidence images \(E\). An MLLM generates initial question--answer pairs based on the visual content of the evidence image. Subsequently, we retrieve alternative images of the same entity from Wikipedia and use them as input images \(I\). Finally, we leverage an LLM together with the caption of the evidence image to rewrite the original question, expressing it in terms of the visual characteristics of the entity under the specific scene depicted in the evidence image. The rewritten question serves as the input question \(Q\).

\paragraph{Internal Visual Reasoning Question Generation.}
We first collect single images containing multiple fine-grained entities from existing datasets such as CelebTo~\cite{zhong2018compact}. In addition, we manually supplement a large number of fine-grained multi-entity scene images through search engines. Based on entity annotations within the images, each entity label is mapped to its corresponding Wikidata~\cite{vrandevcic2014wikidata} identifier through exact string matching followed by manual verification. Using these unique Wikidata identifiers, we retrieve the corresponding Wikipedia pages and establish multimodal links among images, structured Wikidata entries, and image--text Wikipedia pages. Fine-grained entity attributes are extracted from Wikidata through SPARQL queries. An MLLM is then employed to generate textual descriptions of visual characteristics that can uniquely identify each entity within an image. Based on these descriptions, an LLM generates question--answer pairs. For pre-retrieval internal visual reasoning questions, the visual description of an entity is incorporated into the question, and the answer requires retrieving a fine-grained attribute of that entity. For post-retrieval internal visual reasoning questions, a fine-grained attribute of an entity is incorporated into the question, and the answer requires identifying the corresponding visual characteristic of that entity.

\paragraph{Knowledge Base and Evidence Annotation.}
To enable stable and hierarchical evaluation, we construct a multimodal knowledge base consisting of image--text Wikipedia pages extracted from a Wikipedia snapshot containing approximately two million entities. Leveraging the previously established entity--Wikipedia alignments, we annotate the exact sentence or image containing the answer as gold evidence.

\paragraph{Quality Control.}
To ensure dataset quality, we adopt a rigorous quality control process. For each generated \((I,Q,A)\) triplet, we use an MLLM to verify the consistency among the image, question, and answer, retaining only samples with strong semantic alignment. We further remove irrelevant or overly generic entities, questions that are excessively short or contain ambiguous answers, and low-quality images. Through stratified sampling, we balance the distribution of entities, images, attributes, and question types. For all manually annotated components, annotators undergo dedicated training and qualification assessments before participation. As a result, VC-VQA constitutes a high-quality and challenging benchmark for vision-centric multimodal knowledge reasoning.

\subsection{Dataset Statistics}

\begin{table}
\centering
\small
\begin{tabular}{lcr}
      \toprule
      \multicolumn{2}{l}{\textbf{Statistic}} & \textbf{Number} \\
      \midrule
      \multicolumn{2}{l}{(I, Q, A) triplets} & 30,000 \\
      \multicolumn{2}{l}{Ex triplets} & 10,000 \\
      \multicolumn{2}{l}{In-pre triplets} & 10,000 \\
      \multicolumn{2}{l}{In-post triplets} & 10,000 \\
      \midrule
      \multicolumn{2}{l}{Unique I} & 26055 \\
      \multicolumn{2}{l}{Unique Q} & 12395 \\
      \multicolumn{2}{l}{Unique entity} & 5549 \\
      \midrule
      \multicolumn{2}{l}{Average length of Q} & 16.6 \\
      \multicolumn{2}{l}{Average length of A} & 2.9 \\
      \multicolumn{2}{l}{Average \#Evidence} & 1.4 \\
      \bottomrule
    \end{tabular}
\caption{Statistics of VC-VQA.}
\label{dataset_statistics}
\vspace{-5pt}
\end{table}

As shown in Table~\ref{dataset_statistics}, VC-VQA contains 30K \((I,Q,A)\) samples, with 10K samples for each of the three reasoning categories. The dataset includes 26055 unique images and 12395 unique questions, demonstrating substantial diversity and generalization capability. The average lengths of questions and answers are 16.6 and 2.9, respectively. On average, each sample requires 1.4 pieces of evidence for successful reasoning. In addition, Table~\ref{datasets_comparison} provides a comprehensive comparison between VC-VQA and existing knowledge-based VQA benchmarks.

\begin{table*}
  \centering
  \small
\begin{tabular}{lcccc cc}
\toprule

\multirow{2}{*}{\textbf{Model}} & \multicolumn{4}{c}{\textbf{Closed Book}}   & \multicolumn{1}{c}{\textbf{Description}} & \multicolumn{1}{c}{\textbf{Q-Only}} \\
\cmidrule(lr){2-5} \cmidrule(lr){6-6} \cmidrule(lr){7-7} 
& \textbf{Ex} & \textbf{In-pre} & \textbf{In-post} & \textbf{Overall} & \textbf{Overall} & \textbf{Overall} \\
\midrule

\multicolumn{5}{l}{Closed-source Model} \\
GPT-4o~\cite{hurst2024GPT} & 47.5 & 10.4 & 24.0 & 27.3& 10.1 & 2.5 \\
GPT-5.4~\cite{singh2025openai} & 56.6 & 23.8 & 32.0 & 37.5  & 12.5& 2.1\\
Gemini-2.5-Flash~\cite{team2023Gemini} & 42.8 & 41.8 & 45.6 & 43.4  & 8.0& \textbf{5.6}\\
Gemini-3.1-Flash-Lite-preview~\cite{team2023Gemini} & 55.1 & 44.6 & 45.9 & 48.5  & 7.6& 3.0\\
Gemini-3-Flash-preview~\cite{team2023Gemini} & \textbf{58.0} & 57.3 & 53.5 & 56.3  & 9.2& 3.5\\
Gemini-3.1-Pro~\cite{team2023Gemini} & 57.0 & \textbf{60.2} & \textbf{54.1} & \textbf{57.1}  & \textbf{14.7}& 2.2\\
\midrule

\multicolumn{5}{l}{Open-source Model} \\
Qwen2.5-VL-7B-Instruct~\cite{bai2025qwen2} & 52.0 & 39.9 & 28.8 & {40.2}  & 11.2& 8.4\\
Qwen2.5-VL-72B-Instruct~\cite{bai2025qwen2} & 51.3 & 39.8 & 36.4 & 42.5  & 13.3& 5.9\\
Qwen3-VL-8B-Instruct~\cite{bai2025qwen3} & 52.8 & 36.2 & 28.8 & 39.3  & 11.4& 6.3\\
Qwen3-VL-8B-Thinking~\cite{bai2025qwen3} & 51.0 & 29.5 & 31.9 & 37.4  & 10.7& 3.7\\
Qwen3-VL-32B-Instruct~\cite{bai2025qwen3} & 51.9 & 36.5 & 31.3 & 39.9  & 13.5& 7.4\\
Qwen3-VL-32B-Thinking~\cite{bai2025qwen3} & 51.5 & 37.7 & 38.2 & 42.5  & 11.9& 2.8\\
Qwen3.5-9B~\cite{bai2025qwen3} & 50.7 & 34.7 & 40.3 & 41.9  & 10.8& 5.0\\
Qwen3.5-35B-A3B~\cite{bai2025qwen3} & 51.8 & 48.8 & 48.5 & 49.7  & 11.4& 5.6\\
Qwen3.5-27B~\cite{bai2025qwen3} & 52.6 & 44.4 & 44.8 & 47.2  & 13.5& 7.2\\
Qwen3.5-122B-A10B~\cite{bai2025qwen3} & 53.2 & \textbf{57.8} & \textbf{51.7} & \textbf{54.2}  & \textbf{15.0}& \textbf{8.9}\\
Qwen3.6-27B~\cite{bai2025qwen3} & 53.2 & 41.8 & 44.6 & 46.5  & 13.7& 4.8\\
InternVL3\_5-8B~\cite{wang2025internvl3} & 49.5 & 17.8 & 23.7 & 30.3  & 8.5& 6.1\\
MiniCPM-V-4\_5 (8B)~\cite{yu2026minicpm} & 51.0 & 33.5 & 30.5 & 38.4 & 9.6& 6.3 \\
Step3-VL-10B~\cite{huang2026step3} & 49.8 & 32.7 & 34.1 & 38.8 & 9.8& 4.8 \\
GLM-4.1V-9B-Thinking~\cite{hong2025glm} & \textbf{54.0} & 34.2 & 34.3 & 40.8 & 10.2 & 3.4\\

\bottomrule
\end{tabular}
  \caption{Performance comparison under Closed Book, Description and Q-Only settings.}
  \label{model_performance_closed}
\vspace{-5pt}
\end{table*}

\section{Experiments}

\subsection{Evaluation Setup}

We evaluate a diverse set of state-of-the-art multimodal models, including leading proprietary models such as GPT-5.4~\cite{singh2025openai} and Gemini 3.1 Pro~\cite{team2023Gemini}, as well as open-source models such as Qwen3.5-122B-A10B~\cite{bai2025qwen3}. To conduct a comprehensive and robust evaluation, we analyze the impact of model scale by comparing multiple parameter sizes within the Qwen3.5 series (9B, 27B, 35B, and 122B) and the Gemini family (Flash-Lite, Flash, and Pro). Furthermore, we perform a horizontal comparison among representative 7B--10B vision-language models from different architectures and vendors, including Qwen2.5-VL~\cite{bai2025qwen2}, Qwen3-VL~\cite{bai2025qwen3}, Qwen3.5, InternVL3.5~\cite{wang2025internvl3}, MiniCPM-V-4.5~\cite{yu2026minicpm}, Step3-VL~\cite{huang2026step3}, and GLM-4.1V~\cite{hong2025glm}.

We evaluate the models under four settings:
    \textit{(1) \textbf{Closed Book}}: Only the input image and question are provided, without access to external knowledge.
    \textit{(2) \textbf{Oracle}}: Models are additionally provided with manually annotated supporting evidence of varying granularity.
    \textit{(3) \textbf{KB}}: Models are allowed to retrieve information from a controlled knowledge base.
    \textit{(4) \textbf{Web}}: Models can search the open web, which provides realistic but inherently unstable external information.

We adopt accuracy as the primary evaluation metric. Answer correctness is determined using an LLM-as-a-Judge framework. To mitigate potential evaluator bias, the final judgment is obtained through majority voting among GPT-4o~\cite{hurst2024GPT}, Gemini-3-Flash, and Qwen3.6-27B.

\subsection{Closed-Book Results}

As shown in Table~\ref{model_performance_closed}, VC-VQA poses a substantial challenge to current multimodal large language models. Even the strongest proprietary model, Gemini 3.1 Pro, and the strongest open-source model, Qwen3.5-122B-A10B, achieve overall accuracies of only 57.1\% and 54.2\%, respectively. These results indicate that considerable room for improvement remains in vision-centric knowledge question answering. We also observe a consistent performance gain as model size increases. This trend is evident from comparisons across the Qwen3.5 family (9B, 27B, 35B, and 122B) as well as the Gemini family (Flash-Lite, Flash, and Pro), suggesting that scaling remains an effective strategy for improving multimodal knowledge reasoning capabilities.

To further validate the genuinely multimodal nature of VC-VQA, we evaluate models under two ablation settings: \textit{\textbf{Q-Only}}, where only the question is provided, and \textit{\textbf{Description}}, where the input image is replaced with a textual description. As shown in Table~\ref{model_performance_closed}, the average accuracy drops to only 5\% under Q-Only, demonstrating that visual input is indispensable and that models cannot rely solely on language priors or dataset bias. Although the Description setting yields slightly better performance, it still lags behind the \textit{Closed Book} baseline with the full image by 32\%. This substantial gap indicates that textual descriptions cannot fully preserve the rich visual details and semantic information contained in images, further confirming that VC-VQA is a genuinely image-centric benchmark.

\begin{table*}
  \centering
  \small
\begin{tabular}{l cccc cccc c}
\toprule
\multirow{2}{*}{\textbf{Model}} & \multicolumn{4}{c}{\textbf{Entity Name}} & \multicolumn{4}{c}{\textbf{Gold Evidence}} & \textbf{Original} \\
\cmidrule(lr){2-5} \cmidrule(lr){6-9} \cmidrule(lr){10-10}
& \textbf{Ex} & \textbf{In-pre} & \textbf{In-post} & \textbf{Overall} & \textbf{Ex} & \textbf{In-pre} & \textbf{In-post} & \textbf{Overall} & \textbf{Ex} \\
\midrule

\multicolumn{10}{l}{Closed-source Model} \\
GPT-4o & 53.3 & 70.4 & 82.7 & 68.8 & 87.8 & 86.8 & 90.6 & 88.4 & 91.3 \\
GPT-5.4 & \textbf{57.7} & 77.0 & 81.9 & 72.2 & 85.6 & 88.9 & 90.6 & 88.4 & 93.0 \\
Gemini-2.5-Flash & 49.1 & 72.3 & 91.7 & 71.0 & 78.3 & 82.9 & 93.0 & 84.7 & 90.2 \\
Gemini-3.1-Flash-Lite & 56.9 & 87.6 & 83.4 & 76.0 & \textbf{88.4} & \textbf{90.1} & 91.8 & \textbf{90.1} & 93.8 \\
Gemini-3-Flash & 56.6 & 87.9 & 93.1 & 79.2 & 86.4 & 90.0 & 93.2 & 89.9 & 94.2 \\
Gemini-3.1-Pro & 57.5 & \textbf{88.4} & \textbf{93.2} & \textbf{79.7} & 86.4 & 89.9 & \textbf{93.4} & 89.9 & \textbf{94.8} \\
\midrule

\multicolumn{10}{l}{Open-source Model} \\
Qwen2.5-VL-7B-Instruct & 51.8 & 63.7 & 64.9 & 60.1 & 70.6 & 91.0 & 82.1 & 81.2 & 88.0 \\
Qwen2.5-VL-72B-Instruct & 52.1 & 74.1 & 76.8 & 67.6 & 82.3 & 90.8 & 91.4 & 88.2 & 88.7 \\
Qwen3-VL-8B-Instruct & 53.0 & 67.5 & 65.9 & 62.1 & 77.2 & \textbf{91.4} & 86.3 & 85.0 & 90.5 \\
Qwen3-VL-8B-Thinking & 51.4 & 66.7 & 73.1 & 63.7 & 82.5 & 89.1 & 88.7 & 86.7 & 91.2 \\
Qwen3-VL-32B-Instruct & 52.7 & 76.0 & 73.6 & 67.4 & 82.8 & 90.7 & 91.0 & 88.2 & 91.3 \\
Qwen3-VL-32B-Thinking & 52.4 & 76.3 & 83.0 & 70.5 & 84.0 & 89.8 & \textbf{91.7} & 88.5 & 91.4 \\
Qwen3.5-9B & 51.6 & 68.6 & 80.1 & 66.8 & 82.2 & 85.0 & 89.4 & 85.5 & 91.7 \\
Qwen3.5-35B-A3B & 53.6 & 79.6 & 86.5 & 73.2 & 82.2 & 88.9 & 90.6 & 87.2 & 91.9 \\
Qwen3.5-27B & 53.6 & 78.7 & 86.4 & 72.9 & 83.3 & 89.3 & 89.6 & 87.4 & \textbf{93.6} \\
Qwen3.5-122B-A10B & \textbf{55.7} & \textbf{88.2} & \textbf{89.4} & \textbf{77.8} & \textbf{86.7} & 88.4 & 91.3 & \textbf{88.8} & 93.3 \\
Qwen3.6-27B & 53.8 & 76.5 & 84.6 & 71.6 & 81.6 & 89.6 & 89.0 & 86.7 & 92.4 \\
InternVL3\_5-8B & 49.6 & 55.5 & 62.0 & 55.7 & 73.4 & 82.2 & 77.0 & 77.5 & 84.6 \\
MiniCPM-V-4\_5 (8B) & 45.7 & 69.4 & 68.3 & 61.1 & 72.9 & 95.9 & 83.1 & 84.0 & 88.6 \\
Step3-VL-10B & 50.8 & 61.7 & 71.1 & 61.2 & 76.8 & 86.0 & 87.5 & 83.4 & 88.9 \\
GLM-4.1V-9B-Thinking & 54.1 & 72.5 & 79.8 & 68.8 & 84.6 & 90.8 & 90.0 & 88.5 & 91.0 \\

\bottomrule
\end{tabular}
  \caption{Performance comparison under Entity Name, Gold Evidence and Original settings.}
  \label{model_performance_golden}
\end{table*}

\subsection{Oracle Results}

Under the Oracle setting, models are provided with manually annotated evidence, eliminating retrieval errors and revealing the upper bound of multimodal reasoning performance. We consider two levels of evidence:
\textit{(1) \textbf{Gold Evidence}}: The exact supporting sentence from Wikipedia that contains the answer.
\textit{(2) \textbf{Entity Name}}: Only the names of the entities relevant to the image are provided.

As shown in Table~\ref{model_performance_golden}, providing entity names improves performance on internal visual reasoning questions by approximately 39\%. This finding suggests that fine-grained visual entity recognition remains a critical bottleneck for current multimodal models. Once the relevant visual entities are correctly identified, models become substantially more capable of recalling associated textual attribute knowledge. In contrast, performance on external visual reasoning questions improves by only about 1\% when entity names are provided. This result indicates that knowing the entity identity alone offers limited assistance for acquiring fine-grained visual attribute knowledge, which is often encoded directly in images rather than text.

When gold evidence is further provided, overall performance improves by 18\% relative to the \textit{Entity Name} setting. This observation suggests that memorizing and retrieving detailed textual and visual attributes of entities remains a core challenge, leaving significant room for retrieval-enhanced approaches. Furthermore, for external visual reasoning questions, we evaluate models on the \textit{\textbf{Original}} questions $Q_0$ used during data construction, namely, questions that directly query visual attributes in the evidence image itself. The 9\% performance gap between this setting and the \textit{Gold Evidence} setting indicates that multimodal knowledge reasoning is substantially more difficult than pure visual question answering, even when the gold evidence image is available. This gap is likely caused by information loss during cross-image entity alignment and the alignment between textual and visual knowledge representations.

\subsection{KB Results}

Under the KB setting, models are allowed to access a controlled knowledge base, enabling us to evaluate their ability to solve questions through retrieval. We consider two retrieval paradigms for assisting MLLMs: {heuristic retrieval} and {agentic retrieval}.
\textit{(1) \textbf{Heuristic Retrieval}}:
The input image and question are combined as a joint query. The system retrieves the most similar image and its associated caption, and further obtains the corresponding textual paragraph linked to that image. The retrieved image, caption, and text paragraph are then provided as evidence to the MLLM.
\textit{(2) \textbf{Agentic Retrieval}}:
The framework consists of three components: a {planner}, an {executor}, and a {reasoner}. The executor is equipped with three tools: image cropping, image retrieval, and text retrieval. The planner iteratively invokes these tools until sufficient evidence has been collected. The executor performs the requested operations, while the reasoner generates the final answer based on the retrieved visual and textual evidence.

In our experiments, both the heuristic retrieval model and the reasoner in the agentic framework use the evaluated MLLM. The planner is implemented with Qwen3.6-27B. We employ BGE-Large-en-v1.5 as the text embedding model and CLIP-ViT-Large as the image embedding model.

\begin{table}[t]
  \centering
\resizebox{\linewidth}{!}{
  \small
\begin{tabular}{lccc}
\toprule
\textbf{Model} & \textbf{Closed} & \textbf{Heuristic} & \textbf{Agentic} \\
\midrule

\multicolumn{4}{l}{Closed-source Model} \\
GPT-4o & 27.3 & 45.3 & 49.6 \\
GPT-5.4 & 37.5 & 47.4 & 52.2 \\
Gemini-2.5-Flash & 43.4 & 47.7 & 50.2 \\
Gemini-3.1-Flash-Lite & 48.5 & 51.3 & 54.4 \\
Gemini-3-Flash & 56.3 & \textbf{57.4} & 60.2 \\
Gemini-3.1-Pro & \textbf{57.1} & \textbf{57.4} & \textbf{61.9} \\
\midrule

\multicolumn{4}{l}{Open-source Model} \\
Qwen2.5-VL-7B-Instruct & 40.2 & 42.0 & 45.8 \\
Qwen2.5-VL-72B-Instruct & 42.5 & 46.2 & 50.5 \\
Qwen3-VL-8B-Instruct & 39.3 & 42.3 & 44.1 \\
Qwen3-VL-8B-Thinking & 37.4 & 43.2 & 47.1 \\
Qwen3-VL-32B-Instruct & 39.9 & 45.0 & 50.2 \\
Qwen3-VL-32B-Thinking & 42.5 & 45.8 & 50.7 \\
Qwen3.5-9B & 41.9 & 43.8 & 47.2 \\
Qwen3.5-35B-A3B & 49.7 & 51.5 & 54.5 \\
Qwen3.5-27B & 47.2 & 49.7 & 54.3 \\
Qwen3.5-122B-A10B & \textbf{54.2} & \textbf{55.4} & \textbf{58.6} \\
Qwen3.6-27B & 46.5 & 49.0 & 53.5 \\
InternVL3\_5-8B & 30.3 & 39.0 & 43.4 \\
MiniCPM-V-4\_5 (8B) & 38.4 & 41.1 & 45.5 \\
Step3-VL-10B & 38.8 & 40.5 & 45.2 \\
GLM-4.1V-9B-Thinking & 40.8 & 44.6 & 48.4 \\

\bottomrule
\end{tabular}
}
  \caption{Performance comparison under KB setting.}
  \label{model_performance_KB}
\vspace{-10pt}
\end{table}

As shown in Table~\ref{model_performance_KB}, agentic retrieval consistently outperforms heuristic retrieval across all settings. We attribute this advantage to three main factors. First, agentic retrieval decomposes complex questions into simpler sub-problems, reducing the burden of solving the entire retrieval task in a single step. In contrast, heuristic retrieval attempts to retrieve all necessary evidence at once, often resulting in superficially relevant but uninformative evidence. Second, agentic retrieval supports iterative planning and evidence collection, allowing the system to continue retrieving information until sufficient evidence has been gathered. Third, agentic retrieval can distinguish whether a question requires external textual knowledge or external visual knowledge and selectively retrieve the most relevant modality. By contrast, heuristic retrieval always returns both text and image evidence in a single retrieval step, which may introduce irrelevant information and interfere with reasoning.

\subsection{Web Results}

We further evaluate models in a realistic open-web environment. Specifically, we adopt the same agentic framework described in Section~5.4, consisting of a planner, an executor, and a reasoner. We extend the executor with five core tools: image cropping, text search, webpage browsing, image search, and reverse image search.

The experiments utilize APIs from Hugging Face, Serper, and Jina. Both the planner and the reasoner are implemented using Qwen3.6-27B. To better understand the contribution of individual tools, we additionally evaluate variants of the framework with selected tools removed.

\begin{table}
  \centering
\resizebox{\linewidth}{!}{
  \small
\begin{tabular}{lcccc}
\toprule
\textbf{Setting} & \textbf{Ex} & \textbf{In-pre} & \textbf{In-post} & \textbf{Overall} \\
\midrule
Closed Book & 53.2 & 41.8 & 44.6 & 46.5 \\
Gold Evidence & 81.6 & 89.6 & 89.0 & 86.7 \\
\midrule
Agentic KB Retrieve & 60.3 & 50.3 & 49.8 & 53.5 \\
\midrule
Agentic Web Search & 62.2 & 50.7 & 51.5 & 54.8 \\
 - w/o Image Search & 30.8 & 51.3 & 50.1 & 44.1 \\
 - w/o Reverse Image Search & 56.4 & 43.8 & 45.7 & 48.7 \\
\bottomrule
\end{tabular}
}
  \caption{Performance comparison under Web setting.}
  \label{model_performance_web}
  \vspace{-10pt}
\end{table}

As shown in Table~\ref{model_performance_web}, the agentic framework achieves substantial improvements over the closed-book setting in the open-web environment. Its performance even surpasses both heuristic and agentic retrieval under the controlled KB setting. When image search is removed, accuracy on {Ex} questions drops dramatically, while performance on {In} questions remains relatively stable. This observation confirms that {Ex} questions rely heavily on external visual knowledge, which is difficult to acquire from textual sources alone. When reverse image search is removed, accuracy decreases substantially across all question types. This result suggests that the agent relies heavily on reverse image search for entity identification, which serves as a crucial prerequisite for subsequent retrieval and reasoning.

\subsection{Quantitative Analysis}

To better understand model behavior, we analyze the distribution of tool usage under the open-web setting. Specifically, we compute the proportion of calls to five categories of tools across the three task types. This analysis helps diagnose tool preferences under different reasoning paradigms and provides further validation of our task design.

\begin{table}
  \centering
  \small
\begin{tabular}{lccc}
\toprule
\textbf{Tool} & \textbf{Ex} & \textbf{In-pre} & \textbf{In-post} \\
\midrule
Text Search & 26.7 & 29.1 & 33.6 \\
Image Search & 32.2 & 1.9 & 14.2 \\
Reverse Image Search & 12.5 & 21.0 & 8.3 \\
Image Cropping & 6.3 & 26.4 & 16.8 \\
Webpage Browsing & 22.2 & 21.7 & 27.1 \\
\bottomrule
\end{tabular}
  \caption{Proportion of calls to the five tools.}
  \label{tool_comparison}
  \vspace{-10pt}
\end{table}


For {Ex} questions, image search accounts for 32.2\% of all tool invocations, substantially higher than in the other two task categories. This distribution aligns closely with the task design: because textual knowledge alone is insufficient to answer these questions, the model actively seeks external visual evidence through image retrieval. For {In-pre} questions, the usage of image cropping (26.4\%) and reverse image search (21.0\%) increases significantly. This behavior suggests that the model typically first localizes the relevant entity within the input image before proceeding to knowledge retrieval and reasoning. For {In-post} questions, text search (33.6\%) and webpage browsing (27.1\%) dominate the tool usage distribution. This finding indicates that models primarily rely on retrieving and reasoning over large amounts of textual knowledge before returning to the image for verification or grounding.

We recommend reading Appendix~\ref{appx:detail} and~\ref{appx:result} for additional experimental details and results.

\section{Conclusion}

We propose VC-VQA, a large-scale benchmark for vision-centric multimodal knowledge reasoning. VC-VQA explicitly models the role of visual information from different sources (internal vs.\ external images) and at different reasoning stages (pre-retrieval vs.\ post-retrieval), and defines three core sub-tasks accordingly. The dataset is built upon a multimodal knowledge base with traceable gold evidence at both textual and visual levels. We systematically evaluate 21 state-of-the-art multimodal models under four settings: closed-book, oracle, controlled knowledge base retrieval, and open-web search. Experimental results show that current MLLMs still struggle with vision-centric knowledge reasoning, with fine-grained visual knowledge acquisition being the primary bottleneck. Moreover, agent-based planning retrieval consistently outperforms heuristic retrieval, highlighting the importance of multi-step reasoning and cross-modal retrieval. Overall, VC-VQA provides a more systematic, controllable, and challenging benchmark for evaluating visual knowledge reasoning in multimodal models.

\section*{Acknowledgements}

This research is supported by Artificial Intelligence-National Science and Technology Major Project (2023ZD0121200), and the National Natural Science Foundation of China (62437001, 62436001, 62531026), and the Natural Science Foundation of Jiangsu Province under Grant BK20243051, and the Strategic Priority Research Program of Chinese Academy of Sciences under Grant XDB1350103.

\section*{Limitations}

Our benchmark has several limitations. First, although the dataset is constructed through a combination of automatic generation and manual verification, a small number of erroneous or ambiguous samples may still remain. We therefore encourage downstream users to perform additional validation when necessary. Second, the current benchmark is restricted to English and does not support multilingual or cross-lingual evaluation. Extending the benchmark to additional languages and cultural contexts is an important direction for future work. Finally, while we evaluate a wide range of models under diverse experimental settings, we do not systematically investigate the effects of different reasoning strategies. Understanding how reasoning paradigms influence multimodal knowledge reasoning remains an important avenue for future research.

\section*{Ethical considerations}

\subsection*{Data Sources and Licensing}

All images included in our dataset are obtained either from publicly released datasets or from publicly accessible resources with explicit usage permission from the copyright holders. We strictly comply with the licensing terms of all data sources. The dataset is intended solely for academic research and is not intended for commercial use.

\subsection*{Privacy and Personally Identifiable Information}

Images of public figures are collected only from publicly available sources that comply with the usage policies of the original datasets and platforms. We do not collect, publish, or distribute unauthorized private photographs or personally identifiable information.

\subsection*{Bias and Fairness}

Despite efforts to maximize the diversity of images and question--answer pairs, the dataset inevitably inherits cultural and geographical biases from its underlying data sources. We encourage downstream researchers to identify and mitigate these biases when developing and evaluating models. Moreover, the pretrained MLLMs evaluated in this work may inherit linguistic, cultural, and political biases from their training corpora, which are beyond the scope of this study. Experimental results should therefore be interpreted with appropriate caution.

\subsection*{Intended Use and Potential Misuse}

This benchmark is designed to facilitate research on multimodal artificial intelligence, particularly the development and evaluation of vision-centric knowledge reasoning and visual question answering systems. It must not be used for applications that violate individual privacy or enable discriminatory or harmful technologies. Users are expected to comply with applicable ethical guidelines and legal regulations.

\subsection*{Computational Resources and Environmental Impact}

All experiments are conducted using NVIDIA H100 GPUs. Dataset construction and model evaluation require approximately 5,000 GPU hours in total. We employ the {vLLM} framework to accelerate inference and improve GPU memory efficiency. Nevertheless, training and evaluating large language models require substantial computational resources and energy consumption, potentially contributing to carbon emissions. We encourage future researchers to adopt efficient resource scheduling and inference strategies to reduce the environmental impact of large-scale AI research.


\bibliography{custom}

\appendix

\section{Additional Related Work}

\subsection{Visual Question Answering}

Visual Question Answering (VQA) aims to answer natural-language questions based on the visual content of a given image. Early benchmark datasets, including VQA v1/v2~\cite{antol2015vqa, goyal2017making}, DAQUAR~\cite{malinowski2014multi},  CLEVR~\cite{johnson2017clevr}, GQA~\cite{hudson2019gqa}, and FM-IQA~\cite{gao2015you}, focus on reasoning solely from visual cues contained within the image. These benchmarks neither require entity linking nor the retrieval of external knowledge to answer questions.

\subsection{Knowledge-based Visual Question Answering}

Knowledge-based Visual Question Answering (KVQA) extends traditional VQA by requiring models to answer questions involving fine-grained entity attributes using external knowledge. Existing benchmarks, including KB-VQA~\cite{wang2015explicit}, FVQA~\cite{wang2017fvqa}, KVQA~\cite{shah2019kvqa}, S3VQA~\cite{jain2021select}, EVQA~\cite{mensink2023encyclopedic}, ViQuAE~\cite{lerner2022viquae}, DynVQA~\cite{li2024benchmarking}, InfoSeek~\cite{chen2023can}, and M$^3$-VQA~\cite{ma2026m3}, fail to comprehensively model the two key sources of visual information in knowledge reasoning: visual cues from the input image and cross-modal visual evidence retrieved from external knowledge sources. As a result, they provide only a partial characterization of vision-centric knowledge reasoning.

\section{Dataset Construction Details}
\label{appx:dataset}
\subsection{Image Pre-screening}

To construct samples for external visual reasoning, we first select high-quality external evidence images \(E\) from the WIT Wikipedia image--text dataset. The screening pipeline consists of two stages: a heuristic filtering stage followed by fine-grained filtering using an MLLM.

\paragraph{Heuristic Filtering.}
We first apply rule-based string matching to filter candidate image--caption pairs. Specifically, we retain captions containing scene-related prepositions (e.g., {at}, {in}, {on}, {during}, and {for}), indicating that the image describes an entity under a specific scenario rather than only its general attributes. This step efficiently removes images lacking scene-specific visual information.

\paragraph{MLLM-based Filtering.}
The remaining image--caption pairs are then fed into an MLLM with a verification prompt. The model determines whether the sample contains a distinguishable fine-grained entity together with unique scene-specific visual characteristics. Meanwhile, it extracts structured information, including the entity name and the associated scenario. Samples with ambiguous scenes, insufficient visual specificity, or questions answerable solely from text are discarded. The remaining images are used as qualified external evidence \(E\).

\subsection{Entity Linking}

To generate internal visual reasoning questions, image annotations must be linked to Wikidata entities. We adopt a two-stage entity linking procedure consisting of automatic matching followed by manual verification.

In the automatic stage, entity names are queried against Wikidata using the QLever~\cite{bast2017qlever} SPARQL engine. When only one candidate entity is returned, its unique Wikidata identifier (QID) is directly assigned.

If multiple candidates are returned or no valid entity is found, the sample enters manual verification. Annotators resolve entity ambiguities and correct the mappings when necessary. Entities without valid Wikidata entries or corresponding English Wikipedia pages are removed from the dataset.

\subsection{Knowledge Base}

To support stable and challenging retrieval evaluation, we construct a controlled multimodal knowledge base from an English Wikipedia snapshot containing approximately two million entities. We first collect all Wikipedia pages corresponding to entities appearing in our dataset.

To increase retrieval difficulty, we introduce hard negative entities for both text and images. For textual negatives, we encode Wikipedia summaries using {all-mpnet-base-v2} and retrieve the top-10 most similar entities with FAISS based on cosine similarity, from which three are randomly selected as hard negatives. For visual negatives, we extract image embeddings using {CLIP-ViT-Large}, build a FAISS index, retrieve the top-10 visually similar images, and randomly retain three as visual hard negatives. Together with the target entity, these hard negatives form the entity pool of our knowledge base.

For each entity, we further extract fine-grained multimodal evidence from its Wikipedia page. Each image is paired with its caption and the page summary to form an image-based evidence item. For textual evidence, we extract the complete section containing the answer. If that section also contains images, the associated images and captions are preserved to maintain the original multimodal context. To ensure sufficient evidence coverage, entities with fewer than three evidence items are supplemented by additional sections, prioritizing sections containing images. Consequently, each entity is associated with at least three multimodal evidence units in the form of \(\langle\)image, caption, section\(\rangle\).

This design achieves a favorable balance between computational efficiency and evaluation rigor. During evaluation, retrieval is performed over the entire constructed knowledge base rather than a small candidate set. Compared with retrieving from the full two-million-entity Wikipedia corpus, our approach substantially reduces computational cost while preserving retrieval difficulty through carefully designed hard negatives, providing a stable, controlled, and diagnostically informative evaluation environment.

\subsection{Quality Control}

We employ a rigorous quality control pipeline throughout dataset construction.

\paragraph{Annotator Training.}
All annotators hold at least a bachelor's degree and receive systematic training on annotation guidelines and common error cases before participating. Only annotators who pass a qualification assessment are allowed to contribute. Participants are informed of the dataset's intended use and authorship policy before annotation.

\paragraph{Data Filtering.}
We combine automatic and manual quality control procedures. Low-quality images, entities without Wikidata support or corresponding Wikipedia pages, and question--answer pairs lacking valid Wikidata attributes are removed. Questions that are excessively short, excessively long, or factually inconsistent with Wikipedia are discarded. Finally, human annotators review the remaining samples to eliminate ambiguous or poorly formulated questions.

\paragraph{MLLM Verification and Sampling.}
Each candidate \((I,Q,A)\) triplet is further verified by an MLLM, and only samples with consistent image, question, and answer semantics are retained. This process yields approximately one million valid samples. To balance the distributions of entities, images, attributes, and question types, we perform stratified sampling to construct the final benchmark containing 30K samples.

\paragraph{Quality Evaluation.}
To assess annotation quality, domain experts independently review 900 stratified samples randomly selected from the final dataset. Experts infer the answers using the input image, question, Wikipedia page, and supporting evidence, and compare their answers with the dataset annotations. The resulting agreement reaches 98.7\%, outperforming several representative knowledge-based VQA benchmarks, including A-OKVQA (86\%), EVQA (86\%), M$^3$-VQA (93\%), and InfoSeek (95\%). These results demonstrate the high annotation quality and factual reliability of VC-VQA.

\subsection{Bias Avoidance}
During dataset construction, we employ both multimodal large language models (MLLMs) and large language models (LLMs) to assist with data generation and quality control. To mitigate potential model bias arising from the pre-training data or alignment strategy of any single model, we intentionally avoid relying on a single model family. Instead, we leverage a diverse set of state-of-the-art models from different providers, including {Qwen}, {GPT}, and {Gemini}, which are built upon different underlying architectures. This heterogeneous model ensemble not only improves the diversity of the generated samples but also enhances the objectivity, neutrality, and generalizability of the benchmark.

To further examine the potential {self-generation bias}, we conduct a small-scale validation experiment. For a selected evaluated model, we partition its test samples into two subsets: questions generated by the model itself and questions generated exclusively by other models. The results show that the model does not achieve a statistically significant improvement on the questions it generated itself. This finding provides no evidence that models are inherently better at solving their own generated questions, suggesting that our dataset construction process does not introduce noticeable model-specific bias and further supporting the objectivity and fairness of the benchmark.

\section{Experimental Details}
\label{appx:detail}
\subsection{Evaluation Protocol}

\begin{table*}
  \centering
  \small
  \begin{tabular}{lccccccccccc}
\toprule
\textbf{Setting}
&\textbf{Input}
&\textbf{Environment}
&\textbf{Methods} 
&\textbf{Example Models} 
&\textbf{Provided Evidence} 
&\textbf{Knowledge Base} \\
\midrule 

\textit{Closed Book} & \{I, Q\} & -  & End-to-end & Gemini-3.1-Pro & No & - \\

\textit{Oracle} & \{I, Q, E\} & -  & End-to-end & Gemini-3.1-Pro & Yes & - \\

\textit{KB} &  \{I, Q\}& KB & Pipeline   &  Agentic Retrieval & No & Wikipedia \\

\textit{Web} &  \{I, Q\} & Web & Pipeline  & Agentic Search & No & Web \\

\bottomrule
\end{tabular}
\caption{Comparison of different evaluation settings.}
\label{setting}
\end{table*}

\begin{table*}

  \centering
  \small
  \begin{tabular}{lccccccccccc}
\toprule
\textbf{}
& \textbf{Gold Evidence}
& \textbf{Name} 
& \textbf{KB} 
& \textbf{Web} 
& \textbf{Closed Book} 
& \textbf{Description} 
& \textbf{Q-Only} \\
\midrule 

I & \ding{51}\ & \ding{51} & \ding{51} & \ding{51} & \ding{51} & Text & \ding{55} \\

Q & \ding{51} & \ding{51} & \ding{51} & \ding{51} & \ding{51} & \ding{51}& \ding{51} \\

Visual Entity Name & \ding{51} & \ding{51} & Retrieved & Searched & \ding{55} & \ding{55}& \ding{55} \\

Evidence & \ding{51} & \ding{55} & Retrieved & Searched & \ding{55} & \ding{55} & \ding{55}\\

\bottomrule
\end{tabular}
  \caption{Comparison of different benchmarking protocols.}
  \label{setting_comparison}
\end{table*}

We evaluate all models under the settings summarized in Table~\ref{setting} and \ref{setting_comparison}, including Closed-book, Oracle, controlled knowledge base (KB) retrieval, and open-web search.

\subsection{Image Descriptions}
To verify the necessity of visual input, we conduct an ablation study using textual image descriptions. Specifically, we employ Qwen3.5-9B to generate detailed descriptions for every image in the dataset. During evaluation, the original image is replaced with its corresponding description, allowing us to examine whether language alone can capture the visual information required for the task. This setting provides a direct assessment of whether textual representations can substitute for genuine visual signals in multimodal reasoning.

\subsection{Heuristic Retrieval}

We design a heuristic multimodal retrieval pipeline. During query construction, the input image and question are independently encoded into visual and textual embeddings using CLIP-ViT-Large, whose average is taken as the cross-modal query representation. On the knowledge base side, each multimodal evidence item is represented in the same manner by averaging the embeddings of its image and corresponding caption. These fused embeddings are indexed using FAISS to enable efficient dense retrieval.

During inference, the query embedding is matched against the FAISS index to retrieve the most similar evidence items. The retrieval module finally returns the top-5 retrieved images together with their corresponding captions and associated Wikipedia sections, which are provided to the MLLM as supporting evidence.

\subsection{Agentic Retrieval}

To emulate the step-by-step reasoning process of human problem solving, we design an agentic retrieval framework consisting of three components: a Planner, an Executor, and a Reasoner.

The Planner serves as the central controller of the system. It decomposes the original question into a sequence of sub-tasks and continuously determines the next action based on feedback from previous retrieval results. Once sufficient evidence has been collected, the planner terminates the retrieval process and forwards all accumulated evidence to the downstream reasoner.

The Executor carries out the planner's instructions by invoking three built-in tools: an image cropping tool for focusing on local visual regions, a text retrieval tool for acquiring external textual knowledge, and an image retrieval tool for obtaining external visual evidence.

Finally, the Reasoner integrates the retrieved information to produce the final answer. Before reasoning, it removes execution logs and other irrelevant intermediate outputs, retaining only informative multimodal evidence. The final prediction is generated by jointly reasoning over the original input image, the question, and the filtered textual and visual evidence.

\subsection{Text Retrieval}

The text retrieval tool is designed to retrieve relevant textual passages from the knowledge base given a multimodal query consisting of an input image and a natural-language question. We adopt a two-stage cascade retrieval strategy.

In the first stage, {visual localization}, the input image is encoded using {CLIP-ViT-Large}. The resulting visual embedding is searched against a FAISS index constructed over all images in the knowledge base to retrieve the most similar image, thereby identifying the corresponding target entity. Once the entity is determined, the retrieval scope is restricted to the Wikipedia sections associated with that entity.

In the second stage, {text retrieval}, the textual query is encoded using {bge-large-en-v1.5}. Following the recommended retrieval protocol, we prepend the instruction ``Represent this sentence for retrieving relevant passages:'' to every query before encoding. The resulting query embedding is compared with the embeddings of all candidate sections belonging to the identified entity, and the most relevant paragraph is returned as textual evidence.

\subsection{Image Retrieval}

The image retrieval tool retrieves multimodal \(\langle\)image, caption\(\rangle\) evidence from the knowledge base given an input image and question. We employ {CLIP-ViT-Large} as the shared vision--language encoder.

For each evidence item in the knowledge base, we extract the image embedding and the caption embedding, average the two representations, and index the resulting multimodal embedding using FAISS. During retrieval, the input image and question are encoded in the same manner, and their embeddings are averaged to construct a unified multimodal query representation. The query is matched against the FAISS index, and the evidence image with the highest similarity score, together with its associated caption, is returned as the retrieval result.

\subsection{Agentic Web Search}

Under the open-web evaluation setting, we extend the agentic retrieval framework while retaining its three-component architecture consisting of a {Planner}, an {Executor}, and a {Reasoner}. To enable interaction with the real-world Internet, the executor is equipped with five core tools.

\paragraph{Text Search.}
Given a natural-language query, the text search tool invokes a web search engine and returns a ranked list of candidate webpages, including their titles, URLs, and snippets.

\paragraph{Image Search.}
The image search tool retrieves semantically relevant images from the web using a textual query, together with their source webpages, providing external visual evidence that cannot be obtained from text alone.

\paragraph{Reverse Image Search.}
The reverse image search tool uses an input image as the query to retrieve visually similar images from the web and trace them back to their original webpages, facilitating fine-grained entity identification.

\paragraph{Image Cropping.}
The image cropping tool extracts local regions from the input image, allowing the agent to focus on discriminative visual details while suppressing irrelevant background information. The cropped regions can also serve as independent queries for subsequent search.

\paragraph{Webpage Browsing.}
Given a webpage URL, the browsing tool extracts its structured textual content and employs an LLM to summarize lengthy webpages with respect to the current query, producing concise and relevant evidence for downstream reasoning.

\subsection{Backbone Models}

To comprehensively and objectively evaluate the capability boundaries of current multimodal large language models (MLLMs), we select a diverse set of representative models. Our benchmark includes both leading proprietary models, such as {GPT-5.4} and {Gemini-3.1-Pro}, and strong open-source models, including the {Qwen} series, {InternVL3.5}, and {MiniCPM-V-4.5}. The selected models span a wide range of parameter scales, from lightweight 7B--10B models, to medium-scale 27B--35B models, and up to very large models such as {Qwen2.5-VL-72B} and {Qwen3.5-122B-A10B}, allowing us to examine the effect of model scale on performance.

The evaluated models are developed by a diverse set of organizations, including OpenAI, Google, Alibaba, and Shanghai AI Laboratory, reducing potential biases arising from a single training paradigm. In terms of architecture, our benchmark covers not only conventional dense and instruction-tuned models, but also Mixture-of-Experts (MoE) models and recently proposed reasoning-oriented architectures that explicitly emphasize long-chain reasoning (``Thinking'' models), such as {Qwen3-VL-8B-Thinking} and {GLM-4.1V-9B-Thinking}. Finally, while retaining the widely used {GPT-4o} as a baseline, we focus primarily on state-of-the-art models released between the second half of 2025 and early 2026, including {GPT-5.4} (March 2026) and {Qwen3.6-27B} (April 2026), ensuring that our evaluation accurately reflects the latest advances in multimodal large language models.

\section{Evaluation Metrics}
\label{appx:eval}
\subsection{Implementation Details}
We adopt the {LLM-as-a-Judge} paradigm to evaluate model predictions. To mitigate the bias of any individual evaluator, we employ a majority voting strategy that aggregates the judgments of three state-of-the-art LLMs: {GPT-4o}, {Gemini-3-Flash}, and {Qwen-3.6-27B}. The final evaluation result is determined by the majority vote among these three models.

To further improve the efficiency of large-scale evaluation, we introduce an {Exact Match} pre-filter into the automatic evaluation pipeline. Specifically, the model prediction is first compared with the ground-truth answer through exact string matching. If the two are identical, the prediction is directly marked as correct, bypassing the subsequent LLM-based evaluation. Only predictions that differ from the reference answer are forwarded to the {LLM-as-a-Judge} stage for semantic equivalence assessment. This strategy substantially reduces unnecessary LLM inference and API calls while maintaining evaluation quality and accuracy, thereby significantly improving the overall evaluation efficiency.

\subsection{Quality of the Metric}
We find that the three judge models reach unanimous agreement on 97\% of the evaluation samples, demonstrating the reliability of the {LLM-as-a-Judge} paradigm for this task. The majority-voting mechanism further reduces potential biases introduced by any single judge model. Notably, we observe no evidence of {self-preference bias}, where a judge favors its own generations. Moreover, all three judge models produce an identical ranking of the evaluated models, further confirming the objectivity and fairness of our benchmark.

To further validate the reliability of the automatic evaluation framework, we randomly sample 900 model predictions for independent human assessment by domain experts. The experts' judgments agree with the majority-voting results in 99\% of the cases, demonstrating that our evaluation protocol is both reliable and robust.

\section{Additional Experiments}
\label{appx:result}
\subsection{Statistical Analysis}

To ensure the stability and reliability of the experimental results, all reported performance metrics are averaged over three independent runs. In terms of variance, due to the inherent uncontrollability of real-world web environments, the error range of Web-based experiments fluctuates between -1 and 1, while the error ranges for all other controlled settings are within -0.5 and 0.5. In addition, we conduct rigorous statistical analyses on multiple comparative results. The findings show that performance differences across the seven core evaluation settings in our benchmark (Gold Evidence, Name, KB, Web, Closed Book, Description, and Q-Only) are highly statistically significant ($p < 0.01$). Furthermore, pairwise comparisons among Closed Book, heuristic retrieval, and agentic retrieval, as well as the ablation studies under the Web setting, also yield highly significant differences ($p < 0.01$). These significance tests strongly demonstrate the substantial impact of different input conditions and retrieval strategies on cross-modal reasoning capabilities.

\subsection{Additional Results}

\begin{table*}
\centering
\small
\begin{tabular}{lll}
\toprule
\textbf{Organization} & \textbf{Model} & \textbf{Release time} \\
\midrule
\multirow{2}{*}{OpenAI}
& GPT-4o & May 2024 \\
& GPT-5.4 & Mar. 2026 \\
\midrule
\multirow{4}{*}{Google}
& Gemini-2.5-Flash & Jun. 2025 \\
& Gemini-3.1-Flash-Lite-preview & Mar. 2026 \\
& Gemini-3-Flash-preview & Dec. 2025 \\
& Gemini-3.1-Pro & Feb. 2026 \\
\midrule
\multirow{11}{*}{Alibaba}
& Qwen2.5-VL-7B-Instruct & Jan. 2025 \\
& Qwen2.5-VL-72B-Instruct & Jan. 2025 \\
& Qwen3-VL-8B-Instruct & Oct. 2025 \\
& Qwen3-VL-8B-Thinking & Oct. 2025 \\
& Qwen3-VL-32B-Instruct & Oct. 2025 \\
& Qwen3-VL-32B-Thinking & Oct. 2025 \\
& Qwen3.5-9B & Mar. 2026 \\
& Qwen3.5-35B-A3B & Feb. 2026 \\
& Qwen3.5-27B & Feb. 2026 \\
& Qwen3.5-122B-A10B & Feb. 2026 \\
& Qwen3.6-27B & Apr. 2026 \\
\midrule
Shanghai AI Lab & InternVL3\_5-8B & Aug. 2025 \\
OpenBMB & MiniCPM-V-4\_5 (8B) & Aug. 2025 \\
StepFun & Step3-VL-10B & Jan. 2026 \\
Zhipu & GLM-4.1V-9B-Thinking & Jul. 2025 \\
\bottomrule
\end{tabular}
\caption{Model information.}
\label{tab:model_release_time}
\end{table*}

\begin{table*}
  \centering
  \small
\begin{tabular}{l cccc cccc}
\toprule
\multirow{2}{*}{\textbf{Model}} & \multicolumn{4}{c}{\textbf{Heuristic Retrieval}} & \multicolumn{4}{c}{\textbf{Agentic Retrieval}} \\
\cmidrule(lr){2-5} \cmidrule(lr){6-9}
& \textbf{Ex} & \textbf{In-pre} & \textbf{In-post} & \textbf{Overall} & \textbf{Ex} & \textbf{In-pre} & \textbf{In-post} & \textbf{Overall} \\
\midrule

\multicolumn{9}{l}{Closed-source Model} \\
GPT-4o & 55.3 & 42.1 & 38.4 & 45.3 & 59.2 & 48.6 & 40.9 & 49.6 \\
GPT-5.4 & 57.9 & 46.7 & 37.6 & 47.4 & 61.4 & 52.8 & 42.3 & 52.2 \\
Gemini-2.5-Flash & 50.3 & 44.9 & 47.8 & 47.7 & 52.5 & 47.1 & 51.1 & 50.2 \\
Gemini-3.1-Flash-Lite & 57.7 & 47.5 & 48.5 & 51.3 & 60.9 & 49.6 & 52.7 & {54.4} \\
Gemini-3-Flash & 58.6 & 58.5 & {55.1} & 57.4 & 62.4 & 59.9 & 58.4 & 60.2 \\
Gemini-3.1-Pro & 58.1 & 59.1 & 54.9 & {57.4} & 62.8 & 63.4 & {59.6} & 61.9 \\
\midrule

\multicolumn{9}{l}{Open-source Model} \\
Qwen2.5-VL-7B-Instruct & 53.8 & 38.6 & 33.8 & 42.0 & 56.0 & 43.2 & 38.3 & 45.8 \\
Qwen2.5-VL-72B-Instruct & 55.3 & 43.0 & 40.4 & 46.2 & 59.2 & 50.1 & 42.2 & 50.5 \\
Qwen3-VL-8B-Instruct & 54.2 & 39.8 & 33.0 & 42.3 & 56.3 & 40.6 & 35.4 & 44.1 \\
Qwen3-VL-8B-Thinking & 53.4 & 39.1 & 37.1 & 43.2 & 56.8 & 43.1 & 41.5 & 47.1 \\
Qwen3-VL-32B-Instruct & 53.9 & 43.7 & 37.4 & 45.0 & 58.5 & 49.3 & 42.8 & 50.2 \\
Qwen3-VL-32B-Thinking & 53.8 & 43.8 & 39.7 & 45.8 & 59.5 & 49.5 & 43.1 & 50.7 \\
Qwen3.5-9B & 54.2 & 37.6 & 39.5 & 43.8 & 57.6 & 41.1 & 42.9 & 47.2 \\
Qwen3.5-35B-A3B & 54.5 & 48.0 & {52.0} & 51.5 & 57.8 & 51.2 & {54.6} & 54.5 \\
Qwen3.5-27B & 55.8 & 45.7 & 47.6 & 49.7 & 59.9 & 51.5 & 51.6 & 54.3 \\
Qwen3.5-122B-A10B & 57.1 & 58.1 & 51.1 & 55.4 & 61.2 & 61.2 & 53.5 & 58.6 \\
Qwen3.6-27B & 55.3 & 44.9 & 46.9 & {49.0} & 60.3 & 50.3 & 49.8 & {53.5} \\
InternVL3\_5-8B & 52.5 & 36.0 & 28.4 & 39.0 & 56.3 & 40.8 & 33.2 & 43.4 \\
MiniCPM-V-4\_5 & 51.3 & 35.2 & 36.9 & 41.1 & 54.5 & 43.5 & 38.4 & 45.5 \\
Step3-VL-10B & 51.5 & 37.3 & 32.6 & 40.5 & 55.7 & 41.6 & 38.3 & 45.2 \\
GLM-4.1V-9B-Thinking & 56.2 & 38.8 & 38.9 & 44.6 & 59.9 & 45.1 & 40.1 & 48.4 \\

\bottomrule
\end{tabular}
  \caption{Performance comparison between Heuristic Retrieval and Agentic Retrieval models under KB settings.}
  \label{model_performance_retrieval_comparison}
\end{table*}

\begin{table*}
  \centering
  \small
\begin{tabular}{lcccccc}
\toprule
\textbf{Model} & \textbf{Closed Book} & \textbf{KB-Heuristic} & \textbf{KB-Agentic} & \textbf{Name} & \textbf{Gold Evidence} & \textbf{Original} \\
\midrule
GPT-4o & 47.5 & 55.3 & 59.2 & 53.3 & 87.8 & 91.3 \\
GPT-5.4 & 56.6 & 57.9 & 61.4 & {57.7} & 85.6 & 93.0 \\
Gemini-2.5-Flash & 42.8 & 50.3 & 52.5 & 49.1 & 78.3 & 90.2 \\
Gemini-3.1-Flash-Lite & 55.1 & 57.7 & 60.9 & 56.9 & {88.4} & 93.8 \\
Gemini-3-Flash & {58.0} & 58.6 & 62.4 & 56.6 & 86.4 & 94.2 \\
Gemini-3.1-Pro & 57.0 & 58.1 & 62.8 & 57.5 & 86.4 & {94.8} \\
Qwen2.5-VL-7B-Instruct & 52.0 & 53.8 & 56.0 & 51.8 & 70.6 & 88.0 \\
Qwen2.5-VL-72B-Instruct & 51.3 & 55.3 & 59.2 & 52.1 & 82.3 & 88.7 \\
Qwen3-VL-8B-Instruct & 52.8 & 54.2 & 56.3 & 53.0 & 77.2 & 90.5 \\
Qwen3-VL-8B-Thinking & 51.0 & 53.4 & 56.8 & 51.4 & 82.5 & 91.2 \\
Qwen3-VL-32B-Instruct & 51.9 & 53.9 & 58.5 & 52.7 & 82.8 & 91.3 \\
Qwen3-VL-32B-Thinking & 51.5 & 53.8 & 59.5 & 52.4 & 84.0 & 91.4 \\
Qwen3.5-9B & 50.7 & 54.2 & 57.6 & 51.6 & 82.2 & 91.7 \\
Qwen3.5-35B-A3B & 51.8 & 54.5 & 57.8 & 53.6 & 82.2 & 91.9 \\
Qwen3.5-27B & 52.6 & 55.8 & 59.9 & 53.6 & 83.3 & {93.6} \\
Qwen3.5-122B-A10B & 53.2 & 57.1 & 61.2 & {55.7} & {86.7} & 93.3 \\
Qwen3.6-27B & 53.2 & 55.3 & 60.3 & 53.8 & 81.6 & 92.4 \\
InternVL3\_5-8B & 49.5 & 52.5 & 56.3 & 49.6 & 73.4 & 84.6 \\
MiniCPM-V-4\_5 & 51.0 & 51.3 & 54.5 & 45.7 & 72.9 & 88.6 \\
Step3-VL-10B & 49.8 & 51.5 & 55.7 & 50.8 & 76.8 & 88.9 \\
GLM-4.1V-9B-Thinking & {54.0} & 56.2 & 59.9 & 54.1 & 84.6 & 91.0 \\
\bottomrule
\end{tabular}
  \caption{Performance comparison on Ex questions.}
  \label{model_performance_ex}
\end{table*}

\begin{table*}
  \centering
  \small
\begin{tabular}{lccccc}
\toprule
\textbf{Model} & \textbf{Closed Book} & \textbf{KB-Heuristic} & \textbf{KB-Agentic} & \textbf{Name} & \textbf{Gold Evidence} \\
\midrule
GPT-4o & 10.4 & 42.1 & 48.6 & 70.4 & 86.8 \\
GPT-5.4 & 23.8 & 46.7 & 52.8 & 77.0 & 88.9 \\
Gemini-2.5-Flash & 41.8 & 44.9 & 47.1 & 72.3 & 82.9 \\
Gemini-3.1-Flash-Lite & 44.6 & 47.5 & 49.6 & 87.6 & {90.1} \\
Gemini-3-Flash & 57.3 & 58.5 & 59.9 & 87.9 & 90.0 \\
Gemini-3.1-Pro & {60.2} & 59.1 & 63.4 & 88.4 & 89.9 \\
Qwen2.5-VL-7B-Instruct & 39.9 & 38.6 & 43.2 & 63.7 & 91.0 \\
Qwen2.5-VL-72B-Instruct & 39.8 & 43.0 & 50.1 & 74.1 & 90.8 \\
Qwen3-VL-8B-Instruct & 36.2 & 39.8 & 40.6 & 67.5 & {91.4} \\
Qwen3-VL-8B-Thinking & 29.5 & 39.1 & 43.1 & 66.7 & 89.1 \\
Qwen3-VL-32B-Instruct & 36.5 & 43.7 & 49.3 & 76.0 & 90.7 \\
Qwen3-VL-32B-Thinking & 37.7 & 43.8 & 49.5 & 76.3 & 89.8 \\
Qwen3.5-9B & 34.7 & 37.6 & 41.1 & 68.6 & 85.0 \\
Qwen3.5-35B-A3B & 48.8 & 48.0 & 51.2 & 79.6 & 88.9 \\
Qwen3.5-27B & 44.4 & 45.7 & 51.5 & 78.7 & 89.3 \\
Qwen3.5-122B-A10B & {57.8} & 58.1 & 61.2 & {88.2} & 88.4 \\
Qwen3.6-27B & 41.8 & 44.9 & 50.3 & 76.5 & 89.6 \\
InternVL3\_5-8B & 17.8 & 36.0 & 40.8 & 55.5 & 82.2 \\
MiniCPM-V-4\_5 & 33.5 & 35.2 & 43.5 & 69.4 & 95.9 \\
Step3-VL-10B & 32.7 & 37.3 & 41.6 & 61.7 & 86.0 \\
GLM-4.1V-9B-Thinking & 34.2 & 38.8 & 45.1 & 72.5 & 90.8 \\
\bottomrule
\end{tabular}
  \caption{Performance comparison on In-pre questions}
  \label{model_performance_pre}
\end{table*}

\begin{table*}
  \centering
  \small
\begin{tabular}{lccccc}
\toprule
\textbf{Model} & \textbf{Closed Book} & \textbf{KB-Heuristic} & \textbf{KB-Agentic} & \textbf{Name} & \textbf{Gold Evidence} \\
\midrule
GPT-4o & 24.0 & 38.4 & 40.9 & 82.7 & 90.6 \\
GPT-5.4 & 32.0 & 37.6 & 42.3 & 81.9 & 90.6 \\
Gemini-2.5-Flash & 45.6 & 47.8 & 51.1 & 91.7 & 93.0 \\
Gemini-3.1-Flash-Lite & 45.9 & 48.5 & 52.7 & 83.4 & 91.8 \\
Gemini-3-Flash & 53.5 & {55.1} & 58.4 & 93.1 & 93.2 \\
Gemini-3.1-Pro & 54.1 & 54.9 & {59.6} & {93.2} & {93.4} \\
Qwen2.5-VL-7B-Instruct & 28.8 & 33.8 & 38.3 & 64.9 & 82.1 \\
Qwen2.5-VL-72B-Instruct & 36.4 & 40.4 & 42.2 & 76.8 & {91.4} \\
Qwen3-VL-8B-Instruct & 28.8 & 33.0 & 35.4 & 65.9 & 86.3 \\
Qwen3-VL-8B-Thinking & 31.9 & 37.1 & 41.5 & 73.1 & 88.7 \\
Qwen3-VL-32B-Instruct & 31.3 & 37.4 & 42.8 & 73.6 & 91.0 \\
Qwen3-VL-32B-Thinking & 38.2 & 39.7 & 43.1 & 83.0 & 91.7 \\
Qwen3.5-9B & 40.3 & 39.5 & 42.9 & 80.1 & 89.4 \\
Qwen3.5-35B-A3B & 48.5 & {52.0} & {54.6} & 86.5 & 90.6 \\
Qwen3.5-27B & 44.8 & 47.6 & 51.6 & 86.4 & 89.6 \\
Qwen3.5-122B-A10B & 51.7 & 51.1 & 53.5 & {89.4} & 91.3 \\
Qwen3.6-27B & 44.6 & 46.9 & 49.8 & 84.6 & 89.0 \\
InternVL3\_5-8B & 23.7 & 28.4 & 33.2 & 62.0 & 77.0 \\
MiniCPM-V-4\_5 & 30.5 & 36.9 & 38.4 & 68.3 & 83.1 \\
Step3-VL-10B & 34.1 & 32.6 & 38.3 & 71.1 & 87.5 \\
GLM-4.1V-9B-Thinking & 34.3 & 38.9 & 40.1 & 79.8 & 90.0 \\
\bottomrule
\end{tabular}
  \caption{Performance comparison on In-post questions.}
  \label{model_performance_post}
\end{table*}

\begin{table*}
  \centering
  \small
\begin{tabular}{lccccc}
\toprule
\textbf{Model} & \textbf{Closed Book} & \textbf{KB-Heuristic} & \textbf{KB-Agentic} & \textbf{Name} & \textbf{Gold Evidence} \\
\midrule
GPT-4o & 27.3 & 45.3 & 49.6 & 68.8 & 88.4 \\
GPT-5.4 & 37.5 & 47.4 & 52.2 & 72.2 & 88.4 \\
Gemini-2.5-Flash & 43.4 & 47.7 & 50.2 & 71.0 & 84.7 \\
Gemini-3.1-Flash-Lite & 48.5 & 51.3 & 54.4 & 76.0 & {90.1} \\
Gemini-3-Flash & 56.3 & 57.4 & 60.2 & 79.2 & 89.9 \\
Gemini-3.1-Pro & 57.1 & 57.4 & 61.9 & {79.7} & 89.9 \\
Qwen2.5-VL-7B-Instruct & 40.2 & 42.0 & 45.8 & 60.1 & 81.2 \\
Qwen2.5-VL-72B-Instruct & 42.5 & 46.2 & 50.5 & 67.6 & 88.2 \\
Qwen3-VL-8B-Instruct & 39.3 & 42.3 & 44.1 & 62.1 & 85.0 \\
Qwen3-VL-8B-Thinking & 37.4 & 43.2 & 47.1 & 63.7 & 86.7 \\
Qwen3-VL-32B-Instruct & 39.9 & 45.0 & 50.2 & 67.4 & 88.2 \\
Qwen3-VL-32B-Thinking & 42.5 & 45.8 & 50.7 & 70.5 & 88.5 \\
Qwen3.5-9B & 41.9 & 43.8 & 47.2 & 66.8 & 85.5 \\
Qwen3.5-35B-A3B & 49.7 & 51.5 & 54.5 & 73.2 & 87.2 \\
Qwen3.5-27B & 47.2 & 49.7 & 54.3 & 72.9 & 87.4 \\
Qwen3.5-122B-A10B & 54.2 & 55.4 & 58.6 & {77.8} & {88.8} \\
Qwen3.6-27B & 46.5 & 49.0 & 53.5 & 71.6 & 86.7 \\
InternVL3\_5-8B & 30.3 & 39.0 & 43.4 & 55.7 & 77.5 \\
MiniCPM-V-4\_5 & 38.4 & 41.1 & 45.5 & 61.1 & 84.0 \\
Step3-VL-10B & 38.8 & 40.5 & 45.2 & 61.2 & 83.4 \\
GLM-4.1V-9B-Thinking & 40.8 & 44.6 & 48.4 & 68.8 & 88.5 \\
\bottomrule
\end{tabular}
  \caption{Performance comparison on all questions.}
  \label{model_performance_overall}
\end{table*}

Table~\ref{model_performance_retrieval_comparison} presents the detailed performance of all evaluated models under the KB setting for each question category. To further facilitate comparison across different evaluation settings, Tables~\ref{model_performance_ex}, \ref{model_performance_pre}, \ref{model_performance_post} and \ref{model_performance_overall} report the performance of each model on the three task categories and all questions under all experimental settings.

\subsection{Model Scale}

\begin{figure*}
  \centering
  \includegraphics[width=0.8\linewidth]{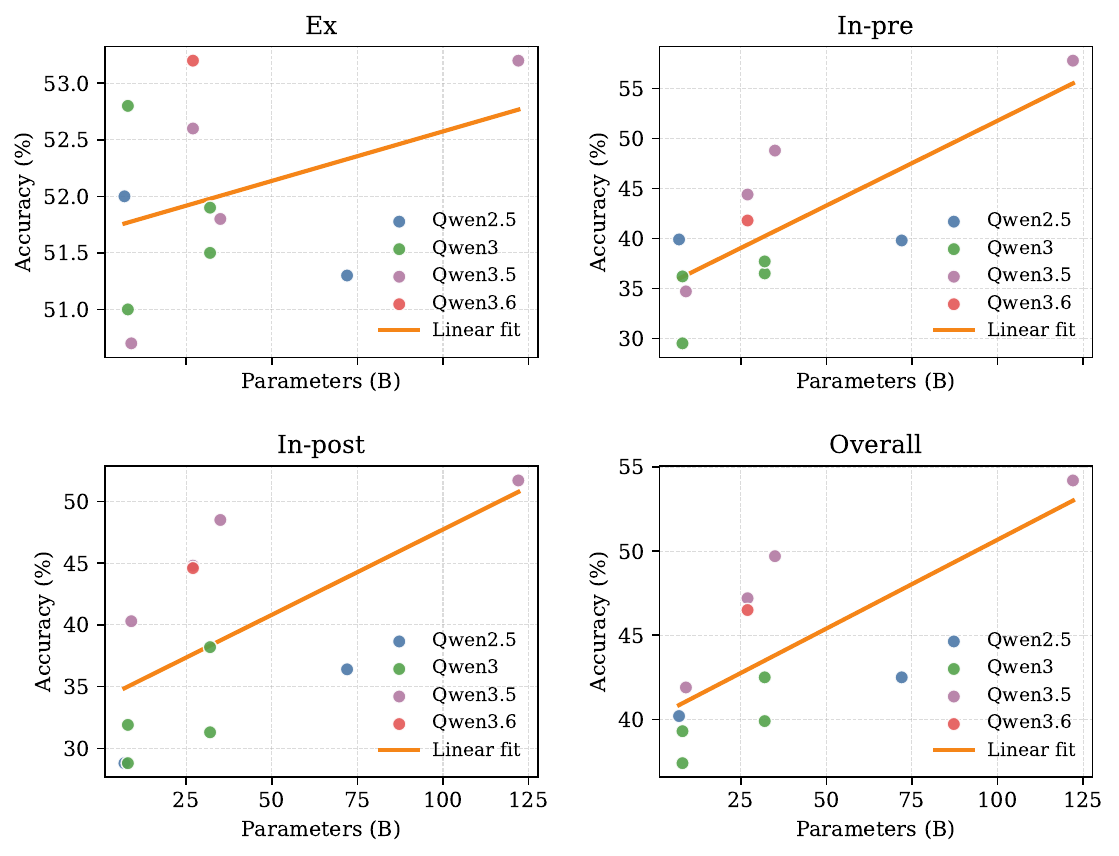}
  \caption{Accuracy of models with different parameter quantities.}
  \label{scale}
\end{figure*}

To further investigate the effect of model scale, we plot the accuracy of the Qwen series against model size for Ex, In-pre, In-post, and overall performance, together with linear regression fits. As shown in Figure~\ref{scale}, performance generally improves as model size increases. However, this scaling effect is much more pronounced on the In-pre and In-post than on the Ex. This suggests that increasing model capacity mainly enhances the memorization of large-scale textual knowledge, while providing only limited benefits for representing and reasoning over external visual knowledge.

\subsection{Retrieval Performance}

To comprehensively evaluate the effectiveness of different retrieval paradigms for complex multimodal queries, we compare heuristic retrieval and agentic retrieval under the controlled knowledge base (KB) setting. As shown in Table~\ref{retrieval_accuracy}, retrieval performance varies substantially across both retrieval methods and task categories.

\begin{table}
  \centering
  \small
\begin{tabular}{lccc}
\toprule
\textbf{Model} & \textbf{Ex} & \textbf{In-pre} & \textbf{In-post} \\
\midrule
Heuristic Retrieval & 38.4	&33.1	&26.5 \\

Agentic Retrieval & 46.5	& 37.7 &	34.6 \\

\bottomrule
\end{tabular}
  \caption{Retrieval accuracy of the heuristic retrieval and agentic retrieval.}
  \label{retrieval_accuracy}
\end{table}

\paragraph{Agentic Retrieval Consistently Outperforms Heuristic Retrieval.}

Agentic retrieval achieves superior performance on all three task categories. Specifically, on external visual reasoning ({Ex}), agentic retrieval achieves a score of 46.5, outperforming heuristic retrieval (38.4) by 8.1 points. Similar improvements are observed for pre-retrieval internal visual reasoning ({In-pre}, 37.7 vs.\ 33.1) and post-retrieval internal visual reasoning ({In-post}, 34.6 vs.\ 26.5).

These results indicate that simple heuristic retrieval based on averaging image and text embeddings is insufficient for fine-grained multimodal knowledge reasoning, as it often loses critical cross-modal semantic information. In contrast, agentic retrieval decomposes complex tasks into multiple reasoning steps, dynamically invokes specialized tools (e.g., image cropping and text retrieval), and iteratively refines retrieval based on intermediate feedback. This enables more accurate understanding of user intent while suppressing irrelevant evidence, resulting in substantially higher retrieval quality.

\paragraph{Hierarchical Difficulty Across Task Types.}

Beyond the choice of retrieval strategy, the intrinsic difficulty of the three reasoning paradigms also leads to a clear performance hierarchy. Regardless of the retrieval method, models achieve the highest performance on {Ex} questions, followed by {In-pre}, while {In-post} remains the most challenging task by a considerable margin.

This trend reflects the increasing complexity of multimodal knowledge reasoning. For {Ex} questions, the retrieval objective is relatively direct, as the required evidence resides in external images. In {In-pre}, models must first identify informative visual cues from the input image before retrieving relevant knowledge, introducing an additional visual reasoning step. The {In-post} setting is substantially more difficult because it requires multi-hop retrieval over multiple candidate entities, followed by cross-modal verification between retrieved textual knowledge and localized image regions. The resulting large search space and high risk of mismatched evidence make post-retrieval reasoning the most challenging scenario for current multimodal retrieval systems.

\subsection{Tool Usage Analysis}

Table~\ref{tool_calls} reports the distribution of tool invocations in the agentic retrieval framework under the KB setting. This analysis reveals the tool preferences of different reasoning paradigms and further validates the design of VC-VQA.

\begin{table}
  \centering
  \small
\begin{tabular}{lccc}
\toprule
\textbf{Tool} & \textbf{Ex} & \textbf{In-pre} & \textbf{In-post} \\
\midrule
Text Retrieval & 17.2 & 62.7 & 68.7 \\
Image Retrieval & 76.4 & 11.0 & 7.5 \\
Image Cropping & 6.3 & 26.4 & 23.8 \\
\bottomrule
\end{tabular}
  \caption{Proportion of calls to the three tools.}
  \label{tool_calls}
\end{table}

For {Ex} questions, image retrieval accounts for 76.4\% of all tool calls. This is expected because these questions require external visual knowledge that cannot be obtained from textual evidence alone, prompting the agent to actively retrieve external images.

For both {In-pre} and {In-post} questions, text retrieval is the most frequently used tool, followed by image cropping. This behavior suggests that the agent typically first localizes the relevant entity in the input image and subsequently retrieves and reasons over textual knowledge associated with that entity.

\subsection{Reasoning Shortcut}

We observe that current MLLMs are prone to a common {reasoning shortcut} when solving post-retrieval internal visual reasoning tasks. When a question asks only about a visual attribute of the target entity, the candidate answer space is often dramatically reduced by the visual context. For example, given the question ``What color clothes is the person born in the United States wearing?'', if the input image contains only three people, the possible answers are effectively limited to the three visible clothing colors. Consequently, a model can often guess the correct answer based solely on visual priors, even without retrieving the required external knowledge.

\begin{table}
  \centering
  \small
\begin{tabular}{lcc}
\toprule
\textbf{Model} & \textbf{Vision only} & \textbf{Name and Vision}  \\
\midrule
Qwen3.5-9B & 75.5 &	40.3 \\
Qwen3.5-27B & 81.4 &	44.8 \\
\bottomrule
\end{tabular}
  \caption{Performance comparison under two settings. }
  \label{shortcut}
\end{table}

As shown in Table~\ref{shortcut}, this shortcut leads to unrealistically high performance under the Closed-book setting. Without requiring entity identification, Qwen3.5-9B and Qwen3.5-27B achieve accuracies of 75.5\% and 81.4\%, respectively.

To eliminate this shortcut, our benchmark requires models to predict both the entity identity and its corresponding visual attribute. For example, instead of asking only for the clothing color, the model must answer ``Who was born in the United States, and where is this person located in the image?''

After introducing the entity identification requirement, model performance drops substantially. Compared with predicting visual attributes alone, the accuracies of Qwen3.5-9B and Qwen3.5-27B decrease by 35.2 and 36.6 percentage points, respectively. Under our evaluation protocol, a prediction is considered correct only if both the entity identity and the visual attribute are correct. This design effectively prevents models from exploiting the limited answer space through guessing and ensures that the benchmark genuinely evaluates the ability to ground externally retrieved knowledge back to the correct visual entity in the input image.

\subsection{The Necessity of External Visual Knowledge}

To further demonstrate the essential role of external visual information in knowledge-intensive question answering, we analyze the problem from three complementary perspectives: static knowledge injection, tool ablation, and model behavior. All three consistently highlight the irreplaceable value of external visual knowledge.

\begin{itemize}
    \item \textbf{Textual knowledge alone is insufficient.} As shown in Section~5.3, providing only the target entity name (\textit{Name}) yields only marginal improvement on questions requiring external visual knowledge. In contrast, when the corresponding evidence image is additionally provided, model performance improves substantially. This result indicates that textual knowledge alone cannot compensate for missing visual details, whereas authentic visual evidence is crucial for accurate reasoning.

    \item \textbf{Removing image retrieval causes a substantial performance drop.} The ablation study in Section~5.5 further supports this conclusion from a dynamic retrieval perspective. When the image search tool is removed from the agent framework, performance on external visual reasoning tasks degrades dramatically. This finding suggests that actively acquiring external visual evidence is not merely beneficial but a necessary component of multimodal reasoning in open-web environments.

    \item \textbf{Models naturally rely on visual search.} As reported in Section~5.6 and Appendix~E.5, models invoke image search significantly more frequently when solving external visual reasoning ({Ex}) questions than other question types. This spontaneous tool-use behavior reflects an intrinsic demand for multimodal evidence and further confirms the necessity of incorporating external visual knowledge into the reasoning process.
\end{itemize}

Overall, whether knowledge is provided directly or acquired through active retrieval, external visual information serves as a fundamental component for enabling genuine multimodal reasoning beyond text-only inference.

\end{document}